\pdfoutput=1
\documentclass[11pt]{article}

\usepackage[preprint]{acl}

\usepackage{times}
\usepackage{latexsym}

\usepackage[T1]{fontenc}
\usepackage[utf8]{inputenc}

\usepackage{microtype}

\usepackage{inconsolata}

\usepackage{graphicx}
\usepackage{amsmath}
\usepackage{multirow}
\usepackage{booktabs}
\usepackage{makecell}
\usepackage{algorithm}
\usepackage{algorithmic}
\usepackage{xcolor}

\usepackage{caption}
\usepackage{url}
\usepackage{amssymb}
\usepackage{pifont}
\usepackage{listings}
\usepackage{tcolorbox}

\definecolor{navy}{RGB}{0, 0, 128}
\definecolor{brickred}{RGB}{203, 65, 84}
\definecolor{darkgreen}{RGB}{0, 100, 0}

\newcommand{\ours}{\textsc{Persona Switch}}
\newcommand{\cmark}{\textcolor{darkgreen}{\ding{51}}}  % green check mark
\newcommand{\xmark}{\textcolor{red}{\ding{55}}}

\title{\ours: Mixing Distinct Perspectives in Decoding Time}
\renewcommand{\thefootnote}{\fnsymbol{footnote}}

\author{%
Junseok Kim$^1$ \quad Nakyeong Yang$^1$ \quad Kyomin Jung$^1$\footnotemark[2]\\
$^1$Seoul National University\\
\texttt{\{kim.junseok,yny0506,kjung\}@snu.ac.kr}
}

\begin{document}
\maketitle
\footnotetext[2]{Corresponding author.} 
\renewcommand{\thefootnote}{\arabic{footnote}}

\begin{abstract}
Role-play prompting is known to steer the behavior of language models by injecting a persona into the prompt, improving their zero-shot reasoning capabilities.
However, such improvements are inconsistent across different tasks or instances.
This inconsistency suggests that zero-shot and role-play prompting may offer complementary strengths rather than one being universally superior.
Building on this insight, we propose \textbf{Persona Switch}, a novel decoding method that dynamically combines the benefits of both prompting strategies.
Our method proceeds step-by-step, selecting the better output between zero-shot and role-play prompting at each step by comparing their output confidence, as measured by the logit gap.
Experiments with widely-used LLMs demonstrate that Persona Switch consistently outperforms competitive baselines, achieving up to 5.13\% accuracy improvement.
Furthermore, we show that output confidence serves as an informative measure for selecting the more reliable output.\footnote{Our code and data have been released on \url{https://github.com/junseokkim00/PersonaSwitch}.}
\end{abstract}

% Large language models (LLMs) can exhibit distinct reasoning behaviors depending on the prompting strategy used, such as zero-shot or role-play prompting.
% While role-play prompting is known to influence model behavior through persona injection, it does not always provide consistent benefits across tasks.
% Motivated by this inconsistency and the complementary strengths of each prompting strategy, we propose a decoding-time framework that dynamically selects the more confident output at each step, guided by the model’s internal confidence signal (logit gap).
% Our method, Confidence-Guided Perspective Selection, generates outputs from multiple prompting perspectives (e.g., zero-shot and role-play), and incrementally builds a response by choosing the more reliable option at each step generation.
% Experiments across five reasoning benchmarks using widely-used LLMs demonstrate that our approach consistently outperforms standard decoding and prompting baselines, achieving up to 5.13\% accuracy improvements.
% Further analysis shows that the logit gap is a robust indicator of output reliability and that our approach benefits from fine-grained token-level selection.
% This decoding-time perspective-switching strategy offers a simple yet effective path to enhance reasoning performance without requiring any additional training.

\section{Introduction}
% Large Language Models (LLMs) have demonstrated remarkable capabilities in solving a wide range of tasks through prompting instructions \citep{brown2020language,kojima2022large}.
% Among various prompting techniques, recent works have explored guiding LLM behavior by assigning personas to the model, known as role-play prompting \cite{shanahan2023role, han2022meet}.
Recent works have explored guiding LLM behavior by assigning personas to the model, known as role-play prompting \citep{shanahan2023role, han2022meet}.
This line of studies has suggested that role-play prompting improves language model performance, such as zero-shot reasoning capability, showing empirical observations reveal that the model's responses can vary depending on the presence of persona \citep{xu2023expertprompting,kong2024better}.
Variations in response suggest that the presence of a persona influences the internal representation of the LLM, potentially due to patterns implicitly learned from the model's training process.

% However, although prior studies have suggested that role-play prompts improve language model performance, subsequent observations indicate that such benefits are not always guaranteed.
However, subsequent observations indicate that role-play prompting does not always guarantee consistent benefits.
% Even when a role-play prompt is prepended to the context, it does not contribute essential information that facilitates problem solving.
This is because prepending a role-play prompt to the context does not contribute essential information that facilitates problem solving.
For instance, consider a simple math problem like \textit{"What is 4 + 8?"}.
% Whether the model receives the question without any persona (zero-shot) or is asked to answer while role-play as a \textit{math teacher}, the key information given to the model is not modified, and the task's complexity is unaffected either.
Whether the model receives the question without any persona (zero-shot) or while role-playing as a \textit{math teacher}, the key information given to the model is unchanged, and the task's complexity is unaffected either.
% Therefore, since a role-play persona does not ensure performance improvement, it is necessary to evaluate whether its inclusion is preferable to its absence.
% Therefore, rather than assuming that a role-play persona universally enhances performance, it is necessary to incorporate a method that can favor its inclusion only when it proves beneficial.
\begin{figure}[t]
\centering
\includegraphics[width=\linewidth]{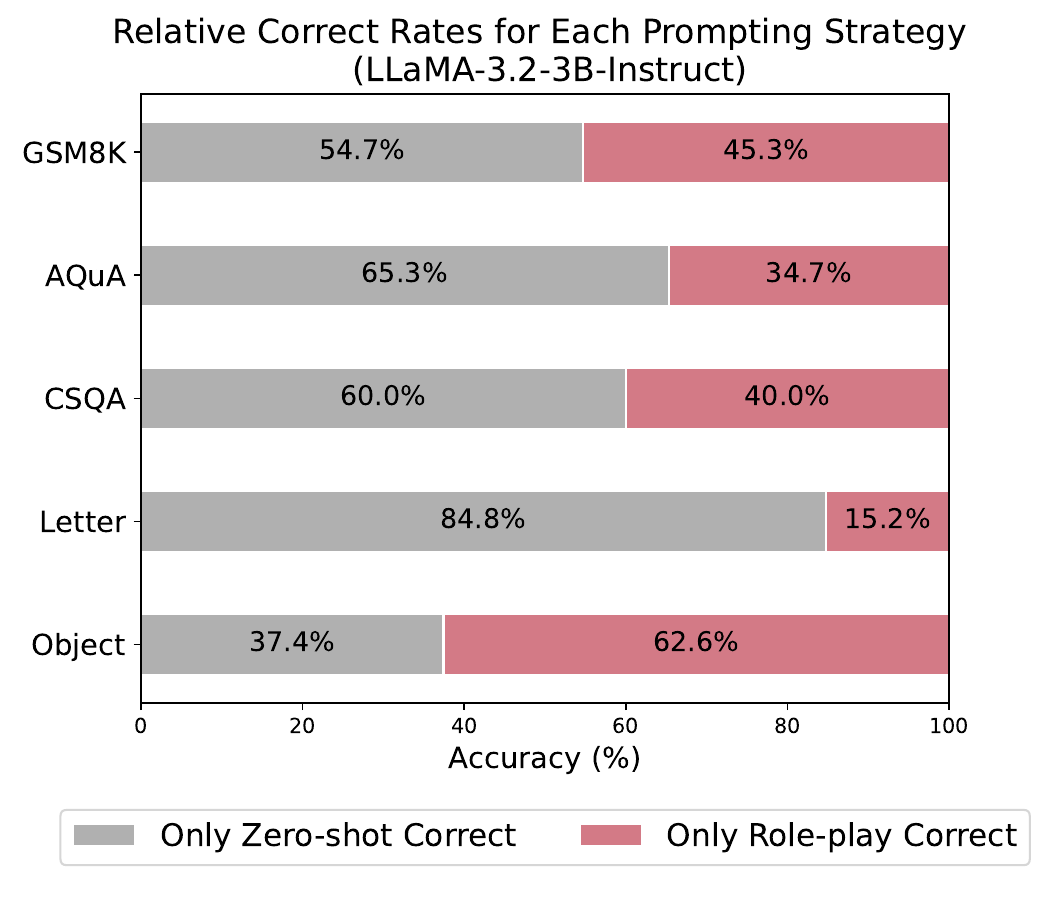}
% \includegraphics[width=\linewidth]{fig_files/ver3.5.pdf}
% \caption{\textbf{Relative Correct Rates Show No Clear Superiority Between Prompting Methods.} Each bar represents the proportion of instances where each method answered correctly, highlighting the relative advantage in method-specific performance.}
\caption{\textbf{Relative Correct Rates Show No Clear Superiority Between Prompting Strategies.} Each bar represents the proportion of instances where only one strategy answered correctly, highlighting that neither one demonstrates clear superiority.}
\label{fig:correctness_distribution}
\vspace{-0.45cm}
\end{figure}
To illustrate this, Figure \ref{fig:correctness_distribution} shows the proportion of cases where only one prompting strategy produces the correct answer.
The result reveals that the superiority depends on each instance, with neither strategy consistently outperforming the other.
% Therefore, rather than assuming that a role-play persona universally enhances performance, it is necessary to identify when its inclusion offers benefits over its absence.
Rather, they possess complementary strengths, underscoring the potential of a unified approach that integrates the strengths of both strategies to achieve better performance.
% Instead, they possess complementary strengths, underscoring the potential of a unified approach that integrates their respective strengths to achieve better performance than conventional prompting strategies.
% Instead, they possess complementary strengths, highlighting the potential of a unified approach that overcomes the limitations of conventional prompting strategies by dynamically integrating their respective advantages.

Motivated by these observations, we propose \textbf{Persona Switch}, a novel decoding method that considers both zero-shot and role-play prompting perspectives.
% Specifically, instead of fully committing to either perspective, our approach generates outputs step-by-step by dynamically selecting the more confident option (based on logit gap) at each step, resulting in responses that blend both perspectives without requiring any additional training.
Specifically, our method compares the single-step outputs generated from different prompting strategies and selects the more reliable one based on the logit gap.
The final response is incrementally constructed by repeating this step-by-step selection process, effectively mixing perspectives of both prompting strategies during decoding.
% To inform this selection, we draw on prior research that shows a model's confidence estimation can serve as a proxy for answer reliability, with low confidence outputs often correlating with hallucinations or incorrect information \cite{zhang2023enhancing}.
% To support this choice, we draw on prior research suggesting that a model's confidence estimation can serve as a proxy for answer reliability, with low confidence outputs often correlating with hallucinations or incorrect information \cite{zhang2023enhancing}.
% Based on this insight, we use the model's internal signal of confidence to estimate the reasoning reliability and choose the higher one between the two outputs.
% Evaluations on five benchmarks demonstrate consistent improvement, achieving 4.42\% of average accuracy gains over the strongest baseline in LLaMA-3.1-8B-Instruct.
Evaluations on five benchmarks demonstrate consistent improvement, achieving 5.13\% of average accuracy gains over the strongest baseline in widely-used LLM.
We also find that the logit gap serves as an informative criterion for selecting reliable outputs, leading to improved performance.
Furthermore, incorporating only informative tokens to assess output confidence enables more fine-grained selection of reliable outputs, further contributing to performance gains.

\section{Related Works}
\paragraph{Role-Play Prompting in LLM} Prompting has become a widely used paradigm for adapting language models to various NLP tasks, with zero-shot, few-shot, and instruction-based prompting, showing strong empirical performance \citep{brown2020language,wei2021finetuned,kojima2022large}.
% Recent works have explored role-play prompting, in which LLMs are guided by persona-based instructions to simulate illustrated behavior \citep{han2022meet, shanahan2023role, xu2023expertprompting, kong2024better}.
Recent works have explored role-play prompting, where LLMs follow persona-based instructions to simulate specific behaviors \citep{han2022meet, shanahan2023role, xu2023expertprompting, kong2024better}.
These approaches report performance gains across specific tasks, though such benefits are not always consistent or well understood \citep{zheng2024helpful, kim2025persona}.
Recent work also highlights that role-play prompting can introduce biases, raising concerns about its robustness and consistent effectiveness across tasks \citep{salewski2023context, gupta2023bias}.
Consequently, the inconsistent gains from role-play prompting suggest that it may complement non-role-play strategies, motivating a unified approach that leverages the strengths of both strategies.

% In parallel, confidence estimation has been studied as a means of improving model reliability, with internal signals such as likelihood scores used to detect incorrect or hallucinated outputs\citep{kadavath2022language, zhang2023enhancing}.
% Our work bridges these approaches by proposing a decoding strategy that selects between zero-shot and role-play outputs based on the model's confidence signals, improving correctness on a per-instance basis.

\section{Methods}
\subsection{Problem Definition}
Role-play prompting is known to effectively steer the model's behavior by inserting personas into the prompt.
However, while adding a persona can change how the model responds, it does not modify the key information available to solve the problem.
We define this invariance as the principle of \textit{Persona-Agnostic Semantic Invariance}.

\paragraph{Persona-Agnostic Semantic Invariance.} 
\textit{Given the same input, introducing a persona does not alter the semantic content of the prompt, thus does not add or remove any task-relevant information.}
\vspace{0.3cm}

\noindent Therefore, if zero-shot and role-play prompting yield different outputs, there is no basis for claiming that either one is inherently superior.
This lack of clear superiority introduces ambiguity when choosing between outputs.
To address this, we propose  \textbf{Persona Switch}, a decoding method that dynamically selects the better output between zero-shot and role-play prompting at each step, based on the model's internal confidence signal (logit gap).

\subsection{Persona Switch}
This section introduces how Persona Switch is executed.
Specifically, Persona Switch consists of two steps: \textbf{Output Candidate Generation} and \textbf{Output Confidence Comparison}.

\paragraph{Output Candidate Generation.}
Given an input $x$, a language model $\mathcal{M}$, and a reasoning trace $\hat{y}^t$ (initially an empty string) at step $t$, we generate an output using greedy decoding from zero-shot $y_\text{base}$ and role-play prompting $y_\text{persona}$. We adopt the role-play prompt $p$ introduced by \citet{kong2024better} (dataset-specific prompts in Appendix \ref{section:RPP}). Following \citet{cao2024step}, we use \texttt{\textbackslash n\textbackslash n} as a stop string to enforce single-step generation. Implementation details about single-step generation are in Appendix \ref{section:output generation}.
The output $y_m~ \scriptstyle(m\in\{\text{base}, \text{persona}\})$ for each prompting strategy is defined as follows:

\begin{table*}[t]
\centering
\renewcommand{\arraystretch}{1.15}                   % vertical breathing room
\small
% \resizebox{\linewidth}{!}{
\scalebox{0.97}{
\begin{tabular}{@{}clccccc|c@{}}
\toprule
\textbf{Model} & \textbf{Method} & \textbf{GSM8K} & \textbf{AQuA} & \textbf{CSQA} & \textbf{Letter} & \textbf{Object} & \textbf{Average} \\
\midrule
\multirow{8}{*}{\makecell{Gemma-2 \\ (2B)}} 
& Greedy      & 63.08 & \underline{31.89} & \underline{67.24} & \underline{11.00} & 31.87 & \underline{41.01} \\
& Top-p   & 61.18 & 28.74 & 66.18 & 6.00 & 29.07 & 38.23 \\
& Top-k   & 59.59 & 28.74 & 66.01 & 4.80 & 30.13 & 37.86 \\
& Multinomial   & 59.21 & 31.10 & 65.77 & 5.20 & \underline{32.27} & 38.71  \\
& Role-Play Prompting & \underline{63.61} & 28.35 & 53.15 & 6.20 & 28.4 & 35.94 \\
\cmidrule{2-8}
& Low-gap Selection       & 61.86 & 31.49 & 57.33 & 5.40  & 31.07 & 37.43 \\
& Random Selection       & 63.25 & 37.93 & 61.73 & 7.86 & 32.44 & 40.64 \\
\cmidrule{2-8}
& \ours       & \textbf{64.14} & \textbf{40.16} & \textbf{67.98} & \textbf{13.60} & \textbf{34.67} & \textbf{44.11 \scriptsize{(+3.10)}} \\
\midrule
\multirow{8}{*}{\makecell{LLaMA-3.2 \\ (3B)}}
% & \multicolumn{7}{l}{\textit{variations}} \\
& Greedy      & \underline{73.09} & 40.94 & \textbf{69.45} & 65.40 & 53.47 & \underline{60.47} \\
& Top-p   & 71.72 & 37.40 & 65.36 & 61.00 & 51.07 & 57.31 \\
& Top-k   & 67.55 & \underline{41.34} & 63.31 & 54.80 & 46.53 & 54.71 \\
& Multinomial   & 73.01 & 38.58 & 64.78 & \underline{66.80} & 51.33 & 58.90 \\
& Role-Play Prompting & 71.49 & 29.53 & 66.99 & 56.00 & \underline{64.13} & 57.63 \\
\cmidrule{2-8}
& Low-gap Selection      & 72.71 & 35.83 & 67.49 & 55.20 & 58.67 & 57.98 \\
& Random Selection       & 74.73 & 40.94 & 68.28 & 62.53 & 61.15 & 61.52 \\
\cmidrule{2-8}
& \ours       & \textbf{76.65} & \textbf{44.88} & \underline{69.12} & \textbf{72.80} & \textbf{64.53} & \textbf{65.60 \scriptsize{(+5.13)}} \\
\midrule
% \multirow{8}{*}{LLaMA-3.1-8B} 
\multirow{8}{*}{\makecell{LLaMA-3.1 \\ (8B)}}
& Greedy      & \underline{83.62} & 51.97 & 71.42 & \underline{82.80} & 58.80 & 69.72 \\
& Top-p   & 79.91 & 54.72 & 68.71 & 72.20 & 64.93 & 68.09 \\
& Top-k   & 79.83 & 51.97 & 68.63 & 66.80 & 61.33 & 65.71 \\
& Multinomial   & 83.17 & \underline{55.12} & 69.62 & 78.00 & 64.67 & \underline{70.12} \\
& Role-Play Prompting & 77.03 & 42.13 & \underline{73.05} & 74.20 & \textbf{71.33} & 67.55 \\
% & \textbf{Ours}       & \textbf{85.75} & \textbf{59.06} & \textbf{73.30} & \textbf{84.20} & \underline{70.40} \\
\cmidrule{2-8}
& Low-gap Selection       & 83.70 & 59.06 & 70.93 & 75.40 & 58.13 & 69.44 \\
& Random Selection       & 84.71 & 58.40 & 72.02 & 77.47 & 64.67 & 71.45 \\
\cmidrule{2-8}
& \ours      & \textbf{85.75} & \textbf{59.06} & \textbf{73.30} & \textbf{84.20} & \underline{70.40} & \textbf{74.54 \scriptsize{(+4.42)}} \\
\bottomrule
\end{tabular}
}
\caption{
\textbf{Experimental results on five reasoning datasets.} We report accuracy (\%) for each method. Bolded numbers represent the best performance for each dataset, while underlined numbers denote the second-best performance. Parentheses indicate Persona Switch’s absolute improvement over the best baseline for each model.
}
\vspace{-0.45cm}
\label{table:main_result}
\end{table*}

\begin{equation}
\begin{aligned}
    y_{\text{base}} = \mathcal{M}(x;\hat{y}^t) \\
    y_{\text{persona}} = \mathcal{M}(p;x;\hat{y}^t)
\end{aligned}
\label{eq:output candidate generation}
\end{equation}

\paragraph{Output Confidence Comparison.} 
This section describes how to select the final output $y^t_{\text{final}}$ for step $t$.
Let $y_m$ denote a sequence of tokens $(y_{m,1}, y_{m,2},\cdots,y_{m,n})$, where $n$ is the length of $y_m$.
We compare their output confidence to select the better one between the two outputs.
% Prior studies have shown that model outputs with low likelihood scores, computed from model logits, often correspond with incorrect or hallucinated outputs, suggesting that output confidence is correlated with response reliability.
This approach is motivated by the observation that low-confidence generations are often correlated with incorrect or hallucinated outputs, as suggested by prior studies that analyzed model confidence estimation and generation quality \citep{kadavath2022language, zhang2023enhancing, wang2024chain}.
Based on this insight, we adopt the logit gap, a token-level confidence measure that computes the difference between the highest and second-highest probabilities for each token.
Formally, the logit gap $\Delta y_{m,j}$ for each token $y_{m, j}$ is computed as follows:
\begin{equation}
\small
    \begin{aligned}
        \Delta y_{m, j} = P(y_{m, j}^1|x,\hat{y}^t,y_{m,<j}) - P(y_{m, j}^2|x,\hat{y}^t,y_{m,<j})
    \end{aligned}
\end{equation}

\noindent where $y_{m,j}^1$ and $y_{m,j}^2$ indicate the top two tokens at the $j$-th decoding step. By summing these logit gaps, we can compute the average logit gap $\Delta y_m$:
\begin{equation}
% \small
    \begin{aligned}
        \Delta y_{m} = \frac{1}{n}\overset{n}{\underset{j=1}{\Sigma}} \Delta y_{m,j}
    \end{aligned}
\end{equation}
After computing $\Delta y_{m}$ for each prompting strategy, we select the output with the larger value as the final output $y^t_{\text{final}}$ for step $t$.
% Then, $y^t_{\text{final}}$ is appended to the reasoning trace $\hat{y}^t$, forming $\hat{y}^{t+1}$.
The reasoning trace is then updated as $\hat{y}^{t+1} = \hat{y}^t \,\|\, y^t_{\text{final}}$ via concatenation.
This process is repeated iteratively until $y^t_{\text{final}}$ is an \textit{eos} token, after which the final answer is extracted from the resulting reasoning trace.
\begin{equation}
    \begin{aligned}
        y^t_{\text{final}} = \begin{cases}
        y_{\text{base}} & \text{if $\Delta{y_\text{base}} > \Delta{y_\text{persona}}$} \\
        y_{\text{persona}} & \text{otherwise}
    \end{cases}
    \end{aligned}
\end{equation}
% \begin{equation}
%     \begin{aligned}
%         \hat{y}^{t+1} = \hat{y}^t \,\|\, y^t_{\text{final}}
%     \end{aligned}
% \end{equation}
\paragraph{Answer Extraction.}
Identifying the exact answer span is challenging.
We follow \citet{kojima2022large} by concatenating an answer trigger prompt to elicit only the final answer.
The answer trigger prompts for each dataset are in Appendix \ref{section:answer extraction}.

\section{Experiments}
% \begin{table*}[h]
% \centering
% \small  % Slightly reduce font size

% % \setlength{\tabcolsep}{10pt}  % Reduce column spacing
% \begin{tabular}{@{}llccccc|c@{}}
% \toprule
% \textbf{Model} & \textbf{Method} & \textbf{GSM8K} & \textbf{AQuA} & \textbf{CSQA} & \textbf{Letter} & \textbf{Object} & \textbf{Average} \\
% \midrule
% \multirow{6}{*}{LLaMA-3.2-3B-Instruct} 
% & Greedy      & \underline{73.09} & 40.94 & \textbf{69.45} & 65.40 & 53.47 & \underline{60.47} \\
% & Top-p   & 71.72 & 37.40 & 65.36 & 61.00 & 51.07 & 57.31 \\
% & Top-k   & 67.55 & \underline{41.34} & 63.31 & 54.80 & 46.53 & 54.71 \\
% & Multinomial   & 73.01 & 38.58 & 64.78 & \underline{66.80} & 51.33 & 58.90 \\
% & Role-Play Prompting & 71.49 & 29.53 & 66.99 & 56.00 & \underline{64.13} & 57.63 \\
% \cmidrule{2-8}
% & \ours       & \textbf{76.65} & \textbf{44.88} & \underline{69.12} & \textbf{72.80} & \textbf{64.53} & \textbf{65.60 \scriptsize{(+5.13)}} \\
% \bottomrule
% \end{tabular}
% \caption{
% \textbf{Full Main results on five reasoning datasets.} We report accuracy (\%) for each method. All zero-shot decoding variants yielded identical results. Bolded numbers represent the best performance for each dataset.
% }
% \vspace{-0.45cm}
% \label{table:qualitative_analysis}
% \end{table*}
\begin{table*}[ht]
\centering
% \small  % Slightly reduce font size
\scriptsize
% \begin{tabular}{@{}lllll@{}}
\begin{tabular}{@{}cp{3.4cm}p{3.5cm}p{3.5cm}p{3.5cm}@{}}
\toprule
\textbf{Dataset} & \textbf{Question} & \textbf{Zero-Shot Response} & \textbf{Role-Play Response} & \textbf{\ours}\\
\midrule
GSM8K & A DVD can be played 1000 times $\cdots$ How many total times can both DVDs be played before they break? & \textcolor{navy}{To find the total number of times both DVDs $\cdots$ So, both DVDs can be played a total of \textbf{2000} times before they break.} \xmark & \textcolor{brickred}{To find the total number of times both DVDs $\cdots$ So, both DVDs can be played a total of \textbf{1509} times before they break.} \cmark & \textcolor{brickred}{To find the total number $\cdots$ The first DVD can $\cdots$} \textcolor{navy}{The second DVD can $\cdots$} \textcolor{brickred}{Therefore, both DVDs can be played a total of \textbf{1509} times before they break.} \cmark \\
\midrule
Letter &  Take the last letters of each words in "Rich Parker Edward Milagros" and concatenate them. & \textcolor{navy}{To find the last letters of each word, $\cdots$ Now, let's concatenate these letters: \textbf{H-R-D-S}.} \cmark & \textcolor{brickred}{To find the last letters of each word, $\cdots$ The resulting word is \textbf{"Hards"}.} \xmark & \textcolor{navy}{To find the last letters of each word, $\cdots$ Now, let's concatenate these letters: \textbf{H-R-D-S}.} \cmark \\
\midrule
Object & Alice, Bob, and Claire are holding a white $\cdots$. Which choice is true? Answer Choices: (A) black ball. (B) green present. (C) orange ball. & \textcolor{navy}{Let's break down the gift swaps: 1. Alice and Bob swap gifts:  $\cdots$ The correct answer is \textbf{(A) black ball.}} \xmark & \textcolor{brickred}{Let's follow the gift exchange process: 1. Alice and Bob swap gifts: $\cdots$ The correct answer is \textbf{(A) black ball.}} \xmark & \textcolor{brickred}{Let's follow the gift exchange process:} \textcolor{navy}{1. Alice and Bob swap gifts:$\cdots$} \textcolor{navy}{The correct answer is \textbf{(B) green present.}} \cmark \\
\bottomrule
\end{tabular}
\caption{
\textbf{Qualitative Analysis on examples of GSM8K, Letter, and Object datasets.} Each example includes responses from zero-shot, role-play prompting, and Persona Switch.
\textcolor{navy}{Blue} indicates response from zero-shot, and \textcolor{brickred}{Red} indicates response from role-play prompting. \cmark and \xmark~indicate correct and wrong, respectively.
}
\vspace{-0.45cm}
\label{table:qualitative_analysis}
\end{table*}
\subsection{Experimental setup}

\paragraph{Experimental Settings.} 
We evaluate our method on five widely-used reasoning benchmarks: GSM8K \cite{cobbe2021training},  AQuA-RAT \cite{ling2017program}, CSQA \cite{talmor2019commonsenseqa}, Last Letter Concatenation \cite{wei2022chain} and Tracking Shuffled-Objects from BIG-bench \cite{srivastava2022beyond}).
% We used the instruction-tuned variants of LLaMA-3 (3.2-3B and 3.1-8B), regarded as sufficiently capable for reasoning tasks \cite{grattafiori2024llama}.
We use two instruction-tuned variants of the LLaMA-3 family, namely LLaMA-3.2-3B-Instruct and LLaMA-3.1-8B-Instruct, and one from Gemma-2 family, Gemma-2-2B-it, following recent studies demonstrating competitive zero-shot reasoning performance \cite{grattafiori2024llama, team2024gemma}.
We compare our method against five baselines (Zero-Shot with Greedy, Top-p, Top-k, Multinomial sampling, and Role-Play Prompting \citep{kong2024better}), as well as two Persona Switch variants: Low-gap Selection (selects the output with lower average logit gap), and Random Selection (randomly selects between two outputs). 
Implementation details are in Appendix \ref{section:Baselines}.

\subsection{Results and analysis}
\paragraph{Main Results.}
As shown in Table \ref{table:main_result}, Persona Switch consistently outperforms competitive baselines.
Our method achieves the highest overall performance regarding average accuracy, surpassing the strongest baseline by 5.13\% and 4.42\%, respectively.
Furthermore, compared to both low-gap selection and random selection, Persona Switch demonstrates superior accuracy across all datasets, underscoring the efficacy of our logit-based answer selection mechanism.
These results validate that selecting responses based on the averaged logit gap yields robust and reliable improvements in multi-domain reasoning tasks.

\paragraph{Analysis of Output Granularity.}
Choosing a proper output unit for our method is crucial, as the final response can vary depending on how often outputs of zero-shot and role-play prompting are compared.
To assess the impact of output granularity, we evaluated our method at the (1) token, (2) step, and (3) full-sequence level using LLaMA-3.2-3B-Instruct.
% Table \ref{table:granularity} shows that step-level selection performs best, indicating that evaluating intermediate reasoning steps provides a more reliable signal of reasoning quality than assessing the whole process, as it effectively incorporates both zero-shot and role-play perspectives.
Table \ref{table:granularity} shows that step-level selection performs best, indicating that evaluating intermediate reasoning steps allows a more diverse integration of zero-shot and role-play perspectives compared to full-sequence selection, leading to improved performance.
On the token level, its evaluation tends to underperform as it requires selecting the correct option for every token, where individual tokens lack sufficient context to distinguish the correct reasoning path.
Additionally, we also compared confidence using maximum logit, and the consistently better performance of logit gap confirms it as a more informative measure.

% \begin{table}[ht]
% \centering
% \small
% \scalebox{0.95}{
% \setlength{\tabcolsep}{3.5pt}
% \begin{tabular}{llcc|cc}
% \toprule
% \multirow{2}{*}{Metric} & \multirow{2}{*}{Granularity} & \multicolumn{2}{c|}{3B} & \multicolumn{2}{c}{8B} \\
% & & GSM8K & Letters & GSM8K & Letters \\
% \midrule
% \multirow{3}{*}{Logit gap} 
% & Token & 75.13 & 71.40 & 85.75 & 81.60 \\
% & Full-Seq & 74.83 & 63.40 & 81.58 & 83.40 \\
% & Step & \textbf{76.65} & \textbf{72.80} & 85.75 & 84.20 \\
% \midrule
% Max logit & Step & 76.27 & 70.60 & \textbf{86.35} & \textbf{86.40} \\
% \bottomrule
% \end{tabular}}
% \caption{
% \textbf{Result for different granularities and different metric on GSM8K and Letters datasets.} Bold indicates the best performance per column.
% }
% \label{table:granularity}
% \vspace{-0.45cm}
% \end{table}
\begin{table}[ht]
\centering
\small
% \scalebox{0.95}{
\resizebox{\linewidth}{!}{
\begin{tabular}{llccc}
\toprule
Metric & Granularity & GSM8K & AQuA & Letters \\
\midrule
Max logit & Step & 76.27 & 42.13 & 70.60 \\
\midrule
\multirow{3}{*}{Logit gap} 
& Token & 75.13 & 41.73 & 71.40 \\
& Full-Seq & 74.83 & 38.58 & 63.40 \\
& \textbf{Step (ours)} & \textbf{76.65} & \textbf{44.88} & \textbf{72.80} \\
\bottomrule
\end{tabular}}
\caption{
\textbf{Performance across granularities and metrics on GSM8K, AQuA, and Letters using LLaMA-3.2-3B-Instruct.} Bold indicates the best performance for each benchmark.
}
\label{table:granularity}
\vspace{-0.45cm}
\end{table}

\paragraph{Analysis of Filtering Stopwords.}
Our approach computes the average logit gap across all tokens when comparing two outputs.
However, not all tokens contribute equally to output semantics, as some carry essential information while others, such as stopwords, do not.
To assess the impact of removing uninformative tokens, we conduct experiments using two filtering strategies: (1) \textbf{stopwords}, excluding stopwords based on NLTK's predefined list\footnote{\url{https://www.nltk.org/howto/corpus.html}}, and (2) \textbf{only numbers}, retaining only numerical values.
Table \ref{table:filtering} shows that applying stopword filtering yields a slight performance gain, suggesting that irrelevant tokens can obscure confidence signals.
In contrast, using only numerical tokens degrades performance, likely due to the loss of linguistically meaningful context.
Findings suggest that the types of meaningful tokens vary across benchmarks, and applying a task-specific filtering strategy can potentially enhance performance.
\begin{table}[h]
\centering
\small
\setlength{\tabcolsep}{3.5pt}
% \scalebox{0.95}{
\resizebox{\linewidth}{!}{
\begin{tabular}{llccc}
\toprule
Model Size & Filtered wordlist & GSM8K & AQuA & Object \\
\midrule
\multirow{3}{*}{3B} & N/A & \textbf{76.65} & 44.88 & 64.53 \\
& only numbers & 76.12 & 40.94 & - \\
& stopwords & 76.27 & \textbf{46.06} & \textbf{64.80} \\
\midrule
\multirow{3}{*}{8B} & N/A & 85.75 & 59.06 & 70.40 \\
& only numbers & 85.52 & 59.84 & - \\
& stopwords & \textbf{85.90} & \textbf{60.24} & \textbf{70.93} \\
\bottomrule
\end{tabular}
}
\caption{
\textbf{Result for different filtered wordlists on GSM8K, AQuA, and Object datasets.} Bold indicates the best performance per column.
}
\label{table:filtering}
\vspace{-0.45cm}
\end{table}

\paragraph{Qualitative Analysis.}
Table \ref{table:qualitative_analysis} presents several examples from three benchmarks using LLaMA-3.1-8B-Instruct.
In the GSM8K and Last Letter examples, we observe that Persona Switch can produce a better answer when one of the two (zero-shot or role-play prompting) is correct.
This demonstrates its ability to identify and favor the more reliable reasoning path.
Furthermore, in the Object Tracking example, even when both zero-shot and role-play responses are incorrect, the Persona Switch successfully produces the correct answer by integrating reasoning steps from both outputs.
These cases illustrate that the Persona Switch enhances model performance by selectively adopting the better step between the two.
Full results for each example can be found in Appendix \ref{section:Full Results}.

\section{Conclusion}
% We present Persona Switch, a dynamic decoding strategy that selects between zero-shot and role-play outputs at each step based on the logit gap.
% By choosing the better response from the two, Persona Switch leverages the strengths of both.
We present Persona Switch, a dynamic decoding strategy that leverages the strengths of both zero-shot and role-play prompting by selecting the better output at each step based on the logit gap.
Notably, it can be easily implemented without extra training and shows performance gains compared to using either prompting strategy alone.
% Ultimately, Persona Switch contributes a simple yet effective decoding-time mechanism that improves reasoning, offering a practical framework for adapting prompting strategies to the specific needs of each instance.
Ultimately, Persona Switch contributes a simple yet effective decoding-time mechanism, offering a practical framework for integrating distinct prompting strategies.

% \newpage
\section*{Limitations}
While our method demonstrates promising results, certain limitations suggest potential directions for future work.
In our experiments, we adopted fixed role-play prompts from prior studies and used their corresponding datasets, which have a clearly defined ground-truth answer.
This setting allowed us to assume persona-agnostic semantic invariance, where the underlying essential information of each input remains consistent regardless of the persona used.
However, in more open-ended or subjective tasks where ground-truth answers are inherently ambiguous, the effectiveness of persona-agnostic approaches may vary, motivating further exploration of a framework suited to such settings.
Furthermore, our evaluation is limited to zero-shot reasoning tasks.
Although our logit-based stepwise selection framework shows strong performance in this context, it remains an open question whether the same approach would be effective in other settings, such as domains beyond reasoning.

% \section{Ethical Considerations}
% \input{texts/ethics}

\section*{Acknowledgements}
This work was supported by Institute of Information \& communications Technology Planning \& Evaluation(IITP) grant funded by the Korea government(MSIT) (No.RS-2023-00229780, Development of Artificial Intelligence Technology for Process-focused Evaluation(Student’s Learning Diagnosis)). K. Jung is with ASRI, Seoul National University, Korea.

\bibliography{custom}
\newpage
\appendix
\section{Dataset-Specific Role-Play Prompts}
\label{section:RPP}
% When utilizing role-play prompting, we adopted dataset-specific role-play prompts that were designed and validated to be effective in \citet{kong2024better}, as shown in Table \ref{table:prompts}.
In our role-play prompting setup, we adopt dataset-specific prompts that were previously designed and validated for task effectiveness by \citet{kong2024better}, as shown in Table \ref{table:prompts}.
Each prompt consists of two messages: 1) \textbf{role-setting prompt}, which assigns a specific persona to the LLM through the user message, and 2) \textbf{role-feedback prompt}, which prompts the model to acknowledge its assigned persona by reinforcing the role through the assistant message.
\begin{table*}[!h]
\centering
\scalebox{0.8}{
\begin{tabular}{l|p{18cm}} 
\toprule
Task &  \multicolumn{1}{c}{Prompt}                                                                                 \\
\midrule
\multirow{2}{*}[-1.4em]{GSM8K} & \textbf{User Message}: From now on, you are an excellent teacher and are teaching your students to get a new word by concatenating the last letters of several words. I am one of your students and want to ask you a related question. \\
\cmidrule{2-2}
& \textbf{Assistant Message}: Of course! I'd be happy to help you with any questions you have about creating new words by concatenating the last letters of several words. Please go ahead and ask your question, and I'll do my best to assist you. \\
\midrule
\multirow{2}{*}[-1.4em]{AQuA} & \textbf{User Message}: From now on, you are an excellent teacher and are teaching your students to get a new word by concatenating the last letters of several words. I am one of your students and want to ask you a related question. \\
\cmidrule{2-2}
& \textbf{Assistant Message}: Of course! I'd be happy to help you with any questions you have about creating new words by concatenating the last letters of several words. Please go ahead and ask your question, and I'll do my best to assist you. \\
\midrule
\multirow{2}{*}[-1.4em]{CSQA} & \textbf{User Message}: From now on, you are a contestant in the general knowledge quiz contest and always answer all kinds of common sense questions accurately. I am the moderator of the game and the final is about to start. \\
\cmidrule{2-2}
 & \textbf{Assistant Message}: That sounds like an exciting challenge! I'm ready to participate in the quiz contest as a contestant. Please go ahead and start the final round—I'm here to provide accurate answers to your common sense questions. \\ 
\midrule
\multirow{2}{*}[-1.4em]{Letter} & \textbf{User Message}: From now on, you are an excellent teacher and are
  teaching your students to get a new word by concatenating the last letters of
  several words. I am one of your students and want to ask you a related
  question.                                                                                                 \\ 
\cmidrule{2-2}
                        & \textbf{Assistant Message}: Of course! I'd be happy to help you with any
  questions you have about creating new words by concatenating the last letters
  of several words. Please go ahead and ask your question, and I'll do my best
  to assist you.                                                                                           \\ 
\midrule
% \multirow{2}{*}[-1.4em]{Coin}   & From now on, you are a coin that always clearly knows
%   which side of your head is facing. Some people want to play a game with you.
%   They may flip you (a coin) or not. And you will tell them if you (a coin) are
%   heads up in the end.                                                                          \\ 
% \cmidrule{2-2}
%                         & Certainly! I'll be your coin for this game. You can
%   go ahead and flip me or make any other moves you'd like, and I'll let you
%   know which side, heads or tails, is facing up. Feel free to start whenever
%   you're ready!                                                                                           \\ 
% \midrule
% \multirow{2}{*}[-1.4em]{Date}   & From now on, you are an excellent teacher and are
%   teaching your students how to calculate dates correctly. I am one of your
%   students and want to ask you a related question.                                                                                                                                     \\ 
% \cmidrule{2-2}
%                         & Of course! I'm here to help you with any questions
%   you have about calculating dates correctly. Please go ahead and ask your
%   question, and I'll do my best to assist you.                                                                                                                                          \\ 
% \midrule
\multirow{2}{*}[-1.4em]{Object} & \textbf{User Message}: From now on, you are a recorder. Alice, Bob, and
  Claire invite you to record a game. They will exchange their stuff in order,
  and you (the recorder) will fully record the whole process and tell them what
  they end up with.                                                                                    \\ 
\cmidrule{2-2}
                        & \textbf{Assistant Message}: Certainly! I will act as a recorder and document the
  game in which Alice, Bob, and Claire will exchange their items. Please
  provide me with the specific order in which they will exchange their
  belongings, and I will keep track of the process and inform you of what each
  person ends up with at the end.  \\
\bottomrule
\end{tabular}}
\caption{Prompts for GSM8K, AQuA, Commonsense Question Answering, Last Letter Concatenation, and Tracking Shuffled Objects. For each task, the upper cell contains the role-setting prompt and the lower cell presents the role-feedback prompt.}
\label{table:prompts}
\end{table*}

\section{Single Step Output Generation}
\label{section:output generation}
To help understand our approach to generating the output as a single step from Output Candidate Generation, we include a Python code example of using stopping criteria and concatenating reasoning trace $\hat{y}$ when inferring. 
More details can be found in the API documentation\footnote{\url{https://huggingface.co/docs/transformers/internal/generation_utils}} of HuggingFace.

\lstset{    
     basicstyle = \scriptsize\ttfamily,             
     numbers = none,                                      
     numberstyle = \footnotesize \color{darkgray},    
     xleftmargin = 0pt,
     commentstyle = \color[RGB]{0,96,96},                 
     stringstyle = \color[RGB]{128,0,0},                  
     frame =tb,                                      
     backgroundcolor = \color{white},         
     language=Python,                                           
     aboveskip = 10pt, 
    belowskip = 10pt
}
\begin{lstlisting}
# A code example of generating output
reasoning_trace = previous outputs

# 1. Apply chat template to the conversation
conversation = [
    {"role": "user", "content": question},
    {"role": "assistant", "content": reasoning_trace}
]
tokenized_chat = tokenizer.apply_chat_template(
    conversation,
    tokenize=False,
    add_generation_prompt=False,
    return_tensors="pt"
)

# 2. Remove the end-of-turn token and encode it
tokenized_chat = tokenized_chat[:-len("<|eot_id|>")]
model_inputs = tokenizer(
    [tokenized_chat],
    return_tensors="pt"
)

# 3. Define stopping criteria
stop_criteria = StoppingCriteria(
    tokenizer, 
    model_inputs['input_ids'].shape(1),
    r"\n\n"
)
stoplist = StoppingCriteriaList([stop_criteria])

# 4. Generate the output with greedy decoding setting
output = model.generate(
    **model_inputs,
    max_new_tokens=256,
    stopping_criteria=stoplist,
    do_sample=False
)
\end{lstlisting}

\section{Answer Extraction}
\label{section:answer extraction}
We extract the final answer by following the method of \citet{kojima2022large}. Specifically, we concatenate the output with the dataset-specific answer trigger prompt, triggering the model to only return the answer. Answer trigger prompts for each dataset could be found in Table \ref{table:answer trigger}.
\begin{table*}[h]
\centering
% \scalebox{0.85}{
\begin{tabular}{ll} 
\toprule
Datasets & Answer
  Trigger                               \\ 
\midrule
GSM8K     & Therefore,
  the answer (arabic numerals) is   \\
AQuA, CSQA    & Therefore,
  among A through E, the answer is  \\
Object    & Therefore,
  among A through C, the answer is  \\
Letter          & Therefore,
  the final answer is               \\
\bottomrule
\end{tabular}%}
\caption{Answer trigger prompts for each dataset.}
\label{table:answer trigger}
\end{table*}

\section{Baselines and Variants}
\label{section:Baselines}
In this section, we introduce the specific configuration for each baseline used in the main result.
We ran the method three times with different seeds (1,2, and 3) and report the average performance.

\paragraph{Greedy decoding}: The model selects the token with the highest probability at each generation. This method is deterministic and often yields fluent but potentially repetitive or overly conservative outputs, as it does not explore alternative candidates.

\paragraph{Top-$p$ sampling (nucleus sampling)}: This method samples the token from the smallest possible set of tokens whose cumulative probability exceeds a predefined threshold $p$. We set $p=0.9$, meaning the model considers a dynamic set of high-probability tokens at each step, encouraging more diverse outputs while maintaining fluency.

\paragraph{Top-$k$ sampling}: The model restricts sampling to the top $k$ most probable tokens at each step, and selects one based on their normalized probabilities. We used $k=10$ in our experiments. This method controls the randomness of generation by limiting the candidate pool, leading to outputs that are both diverse and locally coherent.

\paragraph{Multinomial sampling (temperature sampling)}: This method samples from the full distribution, controlled by a temperature parameter $\tau$. A lower temperature makes the distribution sharper, favoring high-probability tokens, while a higher temperature leads to more randomness. We used $\tau=0.6$, which is the model’s default value.

\paragraph{Low-gap selection}: As a variant of our method, this approach selects outputs based on the average logit gap between zero-shot and role-play prompting outputs for each step. Specifically, it computes the logit difference for each token and selects the output with the lower average gap.

\paragraph{Random selection}: Another variant of our method, this approach randomly selects between the zero-shot and role-play prompting outputs at each step to construct the final response.
This method introduces stochasticity and serves as a baseline for comparison with our method.

\begin{table*}[t]
\centering
\scalebox{0.82}{
\setlength{\tabcolsep}{20pt}
\begin{tabular}{lcccc}
\toprule
Dataset &Answer Format & $N_{q}$ &$L_{q}$ &License \\
\midrule
GSM8K &arabic number &1319 &46.9  &MIT License \\
AQUA-RAT &option (A-E) &254 &51.9  &Apache-2.0 \\
CommonsenseQA &option (A-E) &1221 &27.8  &Unspecified \\
Last Letters &string &500 &15.0  &- \\
Shuffled Objects &option (A-C) &750 &91.1 &Apache-2.0 \\
\bottomrule
\end{tabular}}
\caption{Relevant information of five datasets. $N_{q}$ denotes the number of questions in each dataset. $L_{q}$ denotes the average words of questions in each dataset.}
\label{table:dataset_description}
\end{table*}
\section{Dataset Details}
In this section, we introduce four different domains, including five distinct datasets used for our experiments. Specific details are shown in Table \ref{table:dataset_description}.

\paragraph{Mathematical Reasoning} We use \textbf{GSM8K} \cite{cobbe2021training}, providing math word problems and \textbf{AQuA-RAT} \cite{ling2017program} for multiple choice questions. All questions in this dataset require mathematical reasoning knowledge.

\paragraph{Commonsense Reasoning} We leverage \textbf{CSQA} \cite{talmor2019commonsenseqa}, providing multiple choice questions. This dataset requires reasoning based on commonsense knowledge to solve the problem.

\paragraph{Symbolic Reasoning} We use \textbf{Last Letter Concatenation} \cite{wei2022chain}, which requires concatenating the last letter of the given words in order.

\paragraph{Others} We utilize \textbf{Tracked Shuffling Objects} from Big-Bench \cite{srivastava2022beyond}, which provides a sequence of trading history of the object, and ask for the object's final position.

\section{Impact of Role-Play Prompt design}
\begin{table*}[!h]
\centering
\resizebox{\linewidth}{!}{
\begin{tabular}{c|l|p{18cm}|c} 
\toprule
Dataset & Prompt Type &  \multicolumn{1}{c}{Prompt} & \multicolumn{1}{c}{Accuracy} \\
\midrule
\multirow{12}{*}{AQuA}
& \multirow{1}{*}[-0.7em]{Prompt 1} & \textbf{User Message}: From now on, you are an excellent math teacher and always teach your students math problems correctly. And I am one of your students. & \multirow{1}{*}[-0.7em]{42.91 \%} \\
\cmidrule{2-4}
& \multirow{3}{*}[-1.4em]{Prompt 2} & \textbf{User Message}: From now on, you are an excellent math teacher and always teach your students math problems correctly. And I am one of your students. & \multirow{3}{*}[-1.4em]{42.91 \%} \\
\cmidrule{3-3}
& & \textbf{Assistant Message}: That's great to hear! As your math teacher, I'll do my best to explain mathematical concepts correctly so that you can understand them easily. Feel free to ask any math problems or questions you have, and I'll be glad to assist you. & \\
\cmidrule{2-4}
& \multirow{3}{*}[-1.4em]{Prompt 3 (Ours)} & \textbf{User Message}: From now on, you are an excellent math teacher and always teach your students math problems correctly. And I am one of your students. & \multirow{3}{*}[-1.4em]{\textbf{44.88 \%}} \\
\cmidrule{3-3}
& & \textbf{Assistant Message}: That's great to hear! As your math teacher, I'll do my best to explain mathematical concepts correctly so that you can understand them easily. Feel free to ask any math problems or questions you have, and I'll be glad to assist you. \textbf{Let's dive into the world of mathematics and explore its wonders together!} & \\
\midrule
\midrule
\multirow{10}{*}{Letter}
& \multirow{1}{*}[-0.7em]{Prompt 1} & \textbf{User Message}: From now on, you are an excellent teacher and are teaching your students to get a new word by concatenating the last letters of several words. I am one of your students and want to ask you a related question. & \multirow{1}{*}[-0.7em]{69.8 \%} \\
\cmidrule{2-4}
& \multirow{1}{*}[-1.4em]{Prompt 2} & \textbf{User Message}: From now on, you are an excellent teacher and are teaching your students to get a new word by concatenating the last letters of several words. I am one of your students and want to ask you a related question. & \multirow{1}{*}[-1.4em]{71.2 \%} \\
\cmidrule{3-3}
& & \textbf{Assistant Message}: Of course! Please go ahead and ask your question, and I'll do my best to assist you. & \\
\cmidrule{2-4}
& \multirow{2}{*}[-1.4em]{Prompt 3 (Ours)} & \textbf{User Message}: From now on, you are an excellent teacher and are teaching your students to get a new word by concatenating the last letters of several words. I am one of your students and want to ask you a related question. & \multirow{2}{*}[-1.4em]{\textbf{72.8 \%}} \\
\cmidrule{3-3}
& & \textbf{Assistant Message}: Of course! \textbf{I'd be happy to help you with any questions you have about creating new words by concatenating the last letters of several words.} Please go ahead and ask your question, and I'll do my best to assist you. & \\
\bottomrule
\end{tabular}}
\caption{\textbf{Variants of persona prompts used in our experiments.} Across AQuA and Last Letter, the more detailed Prompt 3 yields the best accuracy, and is therefore adopted as the default in our main experiments.}
\label{table:prompt_design}
\end{table*}
\label{section:impact of prompt design}
We conducted additional experiments to evaluate the performance of Persona Switch using multiple variants of the role-play prompt template (Prompts 1, 2, and 3), which differ in verbosity and structure, following \citet{kong2024better}. Specifically, Prompt 1 consists solely of the user message, Prompt 2 includes a user–assistant message pair, and Prompt 3 extends Prompt 2 with an additional assistant sentence, which serves as the default in our main experiments. As presented in Table \ref{table:prompt_design}, Prompt 3 resulted in the strongest performance across both AQuA and Last Letter tasks. These results suggest that employing a more detailed role-play prompt was associated with improved performance. Based on these findings, we adopted Prompt 3 as the default configuration across all benchmarks.

\section{More ablation studies}
\subsection{Role-Playing vs. Stochastic decoding}
To examine whether the performance gains of Persona Switch can be attributed solely to increased decoding diversity, we replace role-playing prompting with standard sampling-based decoding strategies, including top-$p$, top-$k$, and multinomial sampling, while keeping the rest of the framework unchanged.
As shown in Table~\ref{tab:sampling_ablation}, sampling-based variants consistently underperform Persona Switch on both AQuA and Letter datasets, despite introducing stochastic diversity.
This indicates that Persona Switch benefits not merely from sampling multiple candidates within a single output distribution but from role-playing prompts that induce qualitatively distinct reasoning trajectories by shifting the underlying generation distribution, resulting in more coherent and complementary candidate solutions.
\begin{table}[t]
\centering
\small
\begin{tabular}{lcc}
\toprule
\textbf{Decoding Configuration} & \textbf{AQuA} & \textbf{Letter} \\
\midrule
Greedy + Top-$p$ ($p=0.9$)         & 44.49 & 69.8 \\
Greedy + Top-$k$ ($k=10$)         & 38.58 & 64.4 \\
Greedy + Multinomial ($\tau=0.6$) & 40.16 & 66.6 \\
\midrule
\textbf{Greedy + Role-Play (PersonaSwitch)} & \textbf{44.88} & \textbf{72.8} \\
\bottomrule
\end{tabular}
\caption{Ablation study comparing sampling-based decoding strategies with role-playing prompting in Persona Switch using LLaMA-3.2-3B-Instruct.}
\label{tab:sampling_ablation}
\end{table}

\section{Full Results of Qualitative Analysis}
\label{section:Full Results}
Table \ref{table:example_gsm8k}, \ref{table:example_letter}, and \ref{table:example_object} illustrate the full result for the example GSM8K, Letter, and Object, respectively. Blue and Red colored text indicates zero-shot output and role-play output, respectively.

\section{Use of AI Tools}
During the preparation of this paper, AI tools (e.g., OpenAI's ChatGPT) were used in a limited, supporting capacity. Specifically, they assisted in enhancing the clarity and fluency of the text and in suggesting relevant keywords during the writing process. All conceptual ideas, experimental designs, implementations, analyses, and final interpretations were developed entirely by the authors. The authors independently verified all cited references, and no citation was included solely based on AI-generated content. No private, unpublished, or sensitive information was shared with AI tools beyond what is explicitly described in this paper.
\begin{table*}[h]
\centering
\scalebox{0.8}{
\begin{tcolorbox}

\textbf{Question:} A DVD can be played 1000 times before it breaks. There are two DVDs in the public library, one has been played 356 times and the other has been played 135 times. How many total times can both DVDs be played before they break? \\

\textbf{Zero-Shot:} \textcolor{navy}{To find the total number of times both DVDs can be played before they break, we need to add the number of times each DVD has already been played to the number of times each can be played before breaking, and then add those two totals together. \\\\
DVD 1 has been played 356 times and can be played 1000 times before breaking, so it can be played 1000 - 356 = 644 more times. \\\\
DVD 2 has been played 135 times and can be played 1000 times before breaking, so it can be played 1000 - 135 = 865 more times. \\\\
Now, we add the number of times each DVD can be played before breaking to the number of times each has already been played: \\
356 + 644 = 1000 (for DVD 1) \\
135 + 865 = 1000 (for DVD 2) \\\\
Now we add the two totals together: \\
1000 + 1000 = 2000 \\\\
So, both DVDs can be played a total of \textbf{2000} times before they break.} \xmark \\\noindent\textcolor{gray!60}{\rule{0.95\linewidth}{0.4pt}}\\
\textbf{Role-Play Prompting:} \textcolor{brickred}{To find the total number of times both DVDs can be played before they break, we need to add the number of times each DVD has already been played and then subtract that total from the total number of times each DVD can be played. \\\\
The first DVD can be played 1000 times, and it has already been played 356 times. So, the number of times it can still be played is: \\
1000 - 356 = 644 \\\\
The second DVD can be played 1000 times, and it has already been played 135 times. So, the number of times it can still be played is: \\
1000 - 135 = 865 \\\\
Now, let's add the number of times each DVD can still be played: \\
644 + 865 = 1509 \\\\
So, both DVDs can be played a total of \textbf{1509} times before they break.} \cmark \\\noindent\textcolor{gray!60}{\rule{0.95\linewidth}{0.4pt}}\\
\textbf{\ours:} \textcolor{brickred}{To find the total number of times both DVDs can be played before they break, we need to add the number of times each DVD has already been played and then subtract that total from the total number of times each DVD can be played. \\\\
The first DVD can be played 1000 times, and it has already been played 356 times. So, the number of times it can still be played is: \\ 
1000 - 356 = 644} \\\\
\textcolor{navy}{The second DVD can be played 1000 times, and it has already been played 135 times. So, the number of times it can still be played is: \\
1000 - 135 = 865 \\\\
Now, we add the number of times each DVD can still be played: \\
644 + 865 = 1509} \\\\
\textcolor{brickred}{ Therefore, both DVDs can be played a total of \textbf{1509} times before they break.} \cmark

\end{tcolorbox}}
\caption{Full Example of GSM8K from Qualitative Analysis}
\label{table:example_gsm8k}
\end{table*}
\begin{table*}[h]
\centering
\scalebox{0.8}{
\begin{tcolorbox}

\textbf{Question:}  Take the last letters of each words in "Rich Parker Edward Milagros" and concatenate them. \\

\textbf{Zero-Shot:} \textcolor{navy}{To find the last letters of each word, we need to take the last letter of each word in the given phrase "Rich Parker Edward Milagros". \\\\
1. Rich - H \\
2. Parker - R \\
3. Edward - D \\
4. Milagros - S \\\\
Now, let\'s concatenate these letters: \textbf{H-R-D-S}.} \cmark \\\noindent\textcolor{gray!60}{\rule{0.95\linewidth}{0.4pt}}\\
\textbf{Role-Play Prompting:} \textcolor{brickred}{To find the last letters of each word, we\'ll look at the last letter of each name: \\\\
- Rich: H \\
- Parker: R \\
- Edward: D \\
- Milagros: S \\\\
Now, let's concatenate these letters: H-R-D-S. The resulting word is "\textbf{Hards}".} \xmark \\\noindent\textcolor{gray!60}{\rule{0.95\linewidth}{0.4pt}}\\
\textbf{\ours:} \textcolor{navy}{To find the last letters of each word, we need to take the last letter of each word in the given phrase "Rich Parker Edward Milagros" \\\\
1. Rich - H \\
2. Parker - R \\
3. Edward - D \\
4. Milagros - S \\\\
Now, let's concatenate these letters: \textbf{H-R-D-S}.} \cmark
\end{tcolorbox}}
\caption{Full Example of Letter from Qualitative Analysis}
\label{table:example_letter}
\end{table*}
\begin{table*}[h]
\centering
\scalebox{0.8}{
\begin{tcolorbox}

\textbf{Question:}  Alice, Bob, and Claire are holding a white elephant gift exchange. At the start of the event, they are each holding a present of a different color: Alice has a black ball, Bob has a green present, and Claire has a orange ball. As the event progresses, pairs of people swap gifts. First, Alice and Bob swap their gifts. Then, Bob and Claire swap their gifts. Finally, Claire and Alice swap their gifts. At the end of the event, Claire has the ... Which choice is true ? Answer Choices: (A) black ball. (B) green present. (C) orange ball. \\

\textbf{Zero-Shot:} \textcolor{navy}{"Let's break down the gift swaps: \\\\
1. Alice and Bob swap gifts: \\
   - Alice has the green present (Bob's original gift) \\
   - Bob has the black ball (Alice's original gift) \\\\
2. Bob and Claire swap gifts: \\
   - Bob has the orange ball (Claire's original gift) \\
   - Claire has the green present (Bob's gift from the first swap) \\\\
3. Claire and Alice swap gifts: \\
   - Claire has the black ball (Alice's gift from the first swap) \\
   - Alice has the orange ball (Claire's gift from the second swap) \\\\
At the end of the event, Claire has the black ball.\\\\
The correct answer is \textbf{(A) black ball}.} \xmark \\\noindent\textcolor{gray!60}{\rule{0.95\linewidth}{0.4pt}}\\
\textbf{Role-Play Prompting:} \textcolor{brickred}{Let's follow the gift exchange process: \\\\
1. Alice and Bob swap gifts: \\
   - Alice has the green present (Bob's original gift) \\
   - Bob has the black ball (Alice's original gift) \\
   - Claire still has the orange ball \\\\
2. Bob and Claire swap gifts: \\
   - Bob has the orange ball (Claire's original gift) \\
   - Claire has the green present (Bob's gift after the first swap) \\\\
3. Claire and Alice swap gifts: \\
   - Claire has the black ball (Alice's gift after the first swap) \\
   - Alice has the orange ball (Claire's gift after the second swap) \\\\
At the end of the event, Claire has the black ball. \\\\
The correct answer is \textbf{(A) black ball}.} \xmark \\\noindent\textcolor{gray!60}{\rule{0.95\linewidth}{0.4pt}}\\
\textbf{\ours:} \textcolor{brickred}{Let's follow the gift exchange process:} \\\\
\textcolor{navy}{1. Alice and Bob swap gifts: \\
   - Alice has the green present (Bob's original gift) \\
   - Bob has the black ball (Alice's original gift) \\
   - Claire still has the orange ball (her original gift)} \\\\
\textcolor{brickred}{2. Bob and Claire swap gifts: \\
   - Alice still has the green present \\
   - Bob has the orange ball (Claire's original gift) \\
   - Claire has the black ball (Bob's gift from the first swap)} \\\\
\textcolor{navy}{3. Claire and Alice swap gifts: \\
   - Alice has the black ball (Claire's gift from the second swap) \\
   - Claire has the green present (Alice's gift from the first swap) \\\\
So, at the end of the event, Claire has the green present. \\\\
The correct answer is \textbf{(B) green present}.}
\cmark
\end{tcolorbox}}
\caption{Full Example of Object from Qualitative Analysis}
\label{table:example_object}
\end{table*}

\end{document}